%% file: cvpr21.tex
\documentclass[final]{cvpr}

\usepackage[dvipsnames]{xcolor}
\makeatletter
\@namedef{ver@everyshi.sty}{}
\makeatother
\usepackage{tikz}

\usepackage{times}
\usepackage{epsfig}
\usepackage{graphicx}
\usepackage{amsmath}
\usepackage{amssymb}
\usepackage{rotating}

\usepackage{upgreek}
\usepackage{booktabs}
\usepackage{mathtools}
\usepackage{physics}

\usepackage[pagebackref=true,breaklinks=true,colorlinks,bookmarks=false]{hyperref}

\usepackage{subcaption}

\usepackage[title]{appendix}

\usepackage{adjustbox}
\usepackage{array}
\newcolumntype{R}[2]{%
	>{\adjustbox{angle=#1,lap=\width-(#2)}\bgroup}%
	l%
	<{\egroup}%
}
\newcommand*\rot{\multicolumn{1}{R{65}{0.2em}}}%

\newcommand{\revised}[1]{{{#1}}}

\def\cvprPaperID{5511} %
\def\confYear{CVPR 2021}

\begin{document}

\title{Self-Supervised Collision Handling via Generative 3D Garment Models\\for Virtual Try-On
\vspace{-0.1cm} }
\author{Igor Santesteban$^1$~~~~~~~~~~~Nils Thuerey$^2$~~~~~~~~~~~Miguel A. Otaduy$^1$~~~~~~~~~~~Dan Casas$^1$\\[0.3cm]
$^1$Universidad Rey Juan Carlos, Spain\\[0.1cm]
$^2$Technical University of Munich, Germany\\[0.1cm]
{\tt\small first.last@\{urjc.es\}\{tum.de\}}
}
\twocolumn[{%
	\renewcommand\twocolumn[1][]{#1}%
	\maketitle
	\begin{center}
		\vspace{-0.6cm}
		\includegraphics[width=\linewidth]{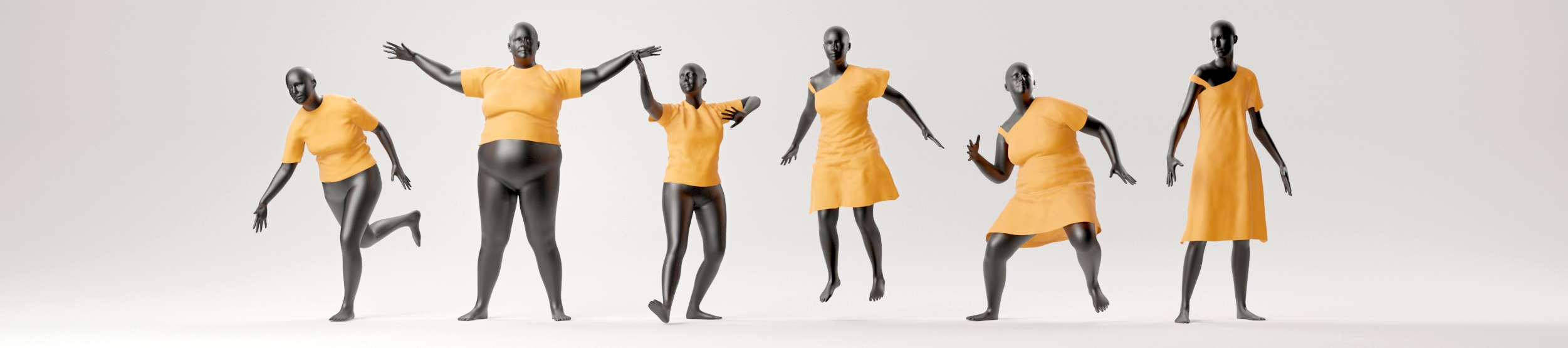}
		\captionof{figure}{
		Our data-driven method regresses deformed garments via a generative model that is trained to avoid collisions. 
		}
		\label{fig:teaser}
	\end{center}
}]

\thispagestyle{empty}

\begin{abstract}
We propose a new generative model for 3D garment deformations that enables us to learn, for the first time, a data-driven method for virtual try-on that effectively addresses garment-body collisions.
In contrast to existing methods that require an undesirable postprocessing step to fix garment-body interpenetrations at test time, our approach directly outputs 3D garment configurations that do not collide with the underlying body.
Key to our success is a new canonical space for garments that removes pose-and-shape deformations already captured by a new diffused human body model, which extrapolates body surface properties such as skinning weights and blendshapes to any 3D point.
We leverage this representation to train a generative model with a novel self-supervised collision term that learns to reliably solve garment-body interpenetrations. 
We extensively evaluate and compare our results with recently proposed data-driven methods, and show that our method is the first to successfully address garment-body contact in unseen body shapes and motions, without compromising realism and detail. 
\end{abstract}

\input{introduction}

\input{related-work}

\input{overview}
\input{method}

\input{regressor}
\input{evaluation}

\input{conclusions}

{\small
\bibliographystyle{ieee_fullname}
\bibliography{cvpr21}
}

\end{document}

%% file: introduction.tex
\section{Introduction} 
The digitalization of 3D garments has important applications in many areas of our everyday lives such as  online shopping, video games, visual effects, and fashion design, and it has traditionally been addressed with physics-based methods \cite{narain2012arcsim,nealen2006physically}.
However, even if these methods offer solutions that generalize well to any type of garment, produce physically-accurate results, and solve body-garment contact, they require computationally expensive runtime evaluations. Consequently, they do not meet the combined robustness and performance needed for real-time applications such as virtual try-on.
Furthermore, they are not easily differentiable and cannot be integrated into computer vision pipelines that, for example, fit deformable models into images to extract information about the scene.

Data-driven methods have emerged as a popular alternative to physics-based methods. The core idea is to \textit{learn} a function that mimics the garment behavior observed in a large dataset.
To this end, recent methods leverage the capability of neural networks to learn nonlinear functions, and propose differentiable models that output 3D deformed garments as a function of the target shape, motion, style, size, and other design parameters~\cite{vidaurre2020fcgnn,santesteban2019virtualtryon,patel2020tailor,tiwari20sizer,wang2018multimodalspace,gundogdu2019garnet,ma2020dressing3d}. 
These methods showcase great realism and robustness, however, we identify a fundamental limitation in all existing works: despite using a loss term that penalizes unphysical body-garment interpenetrations at training time \cite{gundogdu2019garnet,bertiche2020cloth3d}, predicted garments commonly suffer from body-garment interpenetrations in test sequences.
This is usually addressed with an added postprocessing step that pushes the problematic regions of the garment, identified by exhaustive search, outside of the body~\cite{patel2020tailor,santesteban2019virtualtryon}.

The undesired interpenetrations arise from natural residual errors in test samples when optimizing neural networks which, combined with the extremely narrow gap between body surface and garment, can produce artifacts even if the predicted 3D mesh closely matches the ground truth deformed garment.
In this work, we address this inherent limitation and propose, to the best of our knowledge, the first data-driven method to reliably solve garment-body interpenetrations \textit{without} requiring any postprocessing step. We achieve this through three main contributions.

First, we propose to enhance existing human body models \cite{loper2015smpl} by learning to smoothly expand the surface parameters to any 3D point. Intuitively, this allows us to model the deformation at any 3D point, \textit{e.g.}, a vertex of a deformed loose garment, leveraging the deformation capabilities of existing human body models. This expanded human body represents a fundamental building block for our method.

Our second contribution addresses the common assumption, made by existing data-driven models, that garment deformations closely follow the underlying body deformations. 
This popular simplification is often used to define garment models that use skinning parameters based on the closest body vertex in \textit{rest pose}, and subsequently articulate the garment using a standard linear blend skinning (LBS) approach.
We show that simplified transformations to bring ground truth data into a normalized representation, \textit{e.g.} via inverse LBS \cite{santesteban2019virtualtryon,patel2020tailor}, cannot correctly represent the complex deformations that garments exhibit, and often introduce undesirable artifacts.
Instead, we propose a garment model that represents deformations in a novel \textit{unposed} and \textit{deshaped} canonical space by removing deformations already captured via our expanded human body model. 
Since it yields correct skinning attributes for any 3D point, our garment model is designed to not to introduce collisions during projection operations between the canonical space and the posed space. 
 
Our third and most important contribution is to leverage the novel canonical representation of garments to learn a generative subspace of deformations. Garments in this canonical space are encoded with respect to a \textit{constant} reference body configuration. This not only gives an improved representation of garment deformations, but also allows us to reliably learn to solve collisions via self-supervision, by exhaustively sampling the generative space.
We then learn a regressor that outputs mesh deformations encoded in this subspace, and use our garment model to project them to the final deformed state. Since both the deformation subspace and the projection step are designed to avoid collisions, our final deformed 3D garments do not interpenetrate the underlying body mesh, regardless the shape and pose parameter.

%% file: related-work.tex
\section{Related Work}
Current approaches to animate 3D clothing can be classified into: physics-based simulation and data-driven models.

\vspace{-0.2cm} %
\paragraph{Physics-Based Cloth Simulation.} 
Physics-based methods use discretizations of classical mechanics to model how cloth deforms, and typically comprise three steps: computation of internal forces, collision detection, and collision response \cite{nealen2006physically}. 
These methods produce highly-realistic simulations, generalize to different garments, and can handle body-garment collisions, however, 
fail to meet the combined robustness and performance needed for real-time applications such as virtual try-on.

A wide range of strategies have been proposed to address the computational bottleneck in physics-based methods. Recent attempts include approximations of the dynamics to trade physical accuracy for speed~\cite{bender2014survey,bouaziz2014projective,ly2020projective}, adaptive remeshing to refine surface discretization~\cite{lee2010multi,narain2012arcsim}, upsampling details to enrich coarse simulations  \cite{kavan2011upsampling,zurdo2012animating}, and GPU-based solvers~\cite{tang2016cama,fratarcangeli2016vivace,tang2018gpu}.
Another challenge in physics-based simulation is the estimation of the model parameters.
To this end, mechanical approaches have been proposed~\cite{wang2011data,miguel2012data}, but expensive studio settings are required. Alternative methods attempt to recover physical parameters directly from videos by a model fitting process \cite{bhat2003estimating,stoll2010video,mongus2012hybrid} or learn this task directly from data~\cite{bouman2013estimating,wu2016physics,yang2017materialrecovery,rasheed2020friction,runia2020wind}.
Despite the impressive progress towards addressing the critical points in physics-based models, virtual try-on applications require faster and easier to set up methods.

\vspace{-0.2cm} %
\paragraph{Data-Driven Cloth Models.} The deformation of 3D clothing can also be modeled from a data-driven perspective.
Inspired by pioneering works that model pose-dependent surface deformations \cite{lewis2000pose}, many methods propose to \textit{learn} from examples how 3D garments deform as a function of the underlying human body. This has been demonstrated for a variety of goals, including  design \cite{shen2020garmentgeneration,wang2018multimodalspace}, animation ~\cite{huang2020arch,casas2014video4d,wang2019intrisicspace,yang2018analyzing}, and virtual try-on~\cite{guan2012drape,vidaurre2020fcgnn,ma2020dressing3d,patel2020tailor,gundogdu2019garnet,tiwari20sizer}.
Training data in form of 3D meshes can be obtained using physics-based simulations~\cite{bertiche2020cloth3d,narain2012arcsim,li2018implicit}, using 3D reconstruction methods for from either multi-view~\cite{zhang2017detailed,pons2017clothcap,robertini2016model,robertini2017multi} or single image~\cite{alldieck19cvpr,gabeur2019moulding,habermann2020deepcap,onizuka2020tetratsdf,zhu2020deep}, or using hybrid methods~\cite{yu2019simulcap}.

To encode garment deformations a variety of representations have been used, including normal maps~\cite{lahner2018deepwrinkles}, displacement maps~\cite{jin2020pixel}, and geometry images~\cite{pumarola20193dpeople}, but closest to our work are the methods that use 3D positions or offsets to deform a template \cite{patel2020tailor,vidaurre2020fcgnn,santesteban2019virtualtryon,wang2018multimodalspace}.
To parameterize the deformation of clothing, a common strategy is to borrow the parametric space from a human body model \cite{loper2015smpl,joo2018total,xu2020ghum}. However, some methods use geometric features such as PointNet~\cite{qi2017pointnet,gundogdu2019garnet} or 
sketches \cite{wang2018multimodalspace} to describe the target body or garment style instead.
\begin{figure*}
	\centering
	\includegraphics[width=\linewidth]{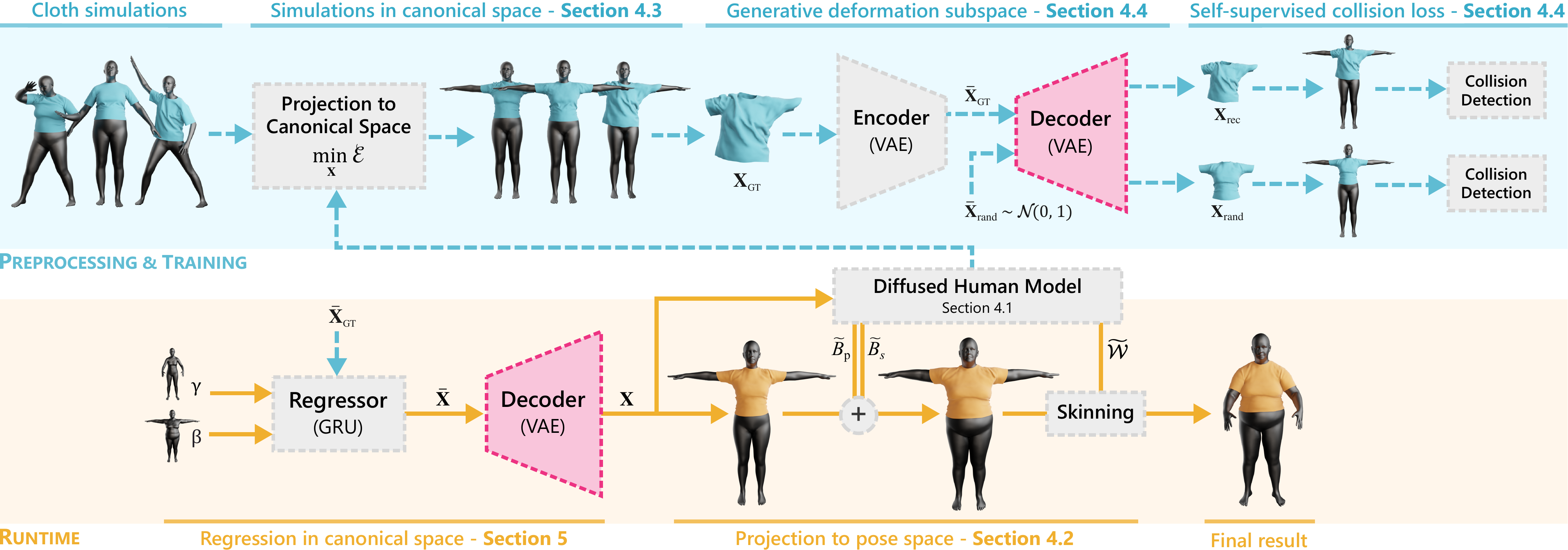}
	\caption{Overview of our preprocessing (top) and runtime pipelines (bottom). The decoder network is trained to avoid collisions in a self-supervised fashion, and then employed by the regressor network to reproduce these states at runtime. 
	}
	\label{fig:overview}
\end{figure*}

Related to our data-driven approach for implementing the regressor, the early work of Guan \etal~\cite{guan2012drape} use a linear model to produce garment deformations due to body pose and shape. The method outputs plausible pose-dependent wrinkles, but shape-dependent deformations are limited to resizing the cloth model.
More recent methods use deep learning to train more sophisticated nonlinear regressors that deform a garment template.
Santesteban \etal~
\cite{santesteban2019virtualtryon} use a fully-connected architecture with GRU \cite{cho2014gru} modules to learn pose and shape garment deformations, including dynamics. 
Similarly, Patel \etal~\cite{patel2020tailor} learn deformations to produce  wrinkles of different scales with a desired style, but are limited to static deformations.
Vidaurre \etal~\cite{vidaurre2020fcgnn} use fully convolutional graph architecture to learn shape-dependent deformations for any garment mesh topology, but do not model pose or style. 
Instead of garment templates, Bertiche \etal~\cite{bertiche2020cloth3d} are capable of learning deformations for a much wider variety of garments using a representation based on 3D offsets 
of the body vertices.
Despite the success of these methods in modeling how cloth deforms as a function of different factors including body shape, pose, motion, design, and material, a common critical weakness in all data-driven approaches is the handling of body-garment collisions. Most methods \cite{patel2020tailor,gundogdu2019garnet,wang2018multimodalspace} use losses that penalize
geometric garment-body penetrations at training time, but all methods require exhaustive postprocess steps to fix collisions in test sequences. We address this remaining challenge with a self-supervised loss enabled by our generative space of garment deformations.

Our work is also related to the recent trend of implicit function learning to predict binary occupancy~\cite{mescheder2019occupancy,saito2019pifu,genova2019structuredimplicit}, signed distance to the surface~\cite{park2019deepsdf,atzmon2020sal,chen2019learning} for any 3D point, or screen space RGB color given camera parameters \cite{sitzmann2019scene,mildenhall2020nerf}. 
To the best of our knowledge, these methods have only been used to encode 3D surfaces, learn generative spaces for shapes, or neural rendering applications.
In contrast, we propose two new uses for implicit function learning: to efficiently solve garment-body collisions with a novel fully differentiable loss term; and to smoothly diffuse 3D shape correctives and rigging weights to any 3D point.

%% file: overview.tex
\section{Method Overview}
Our goal is to learn a function to predict how a 3D garment dynamically deforms given a target human body pose and shape. 
In contrast to existing methods %
\cite{gundogdu2019garnet,patel2020tailor,santesteban2019virtualtryon,bertiche2020cloth3d}, we put special emphasis in learning a model that directly outputs garment geometry that does not interpenetrate with the underlying human body, \textit{i.e.,} it is physically correct after inference without requiring any post-processing. Hence, the final state is not compromised in terms of the regressed garment details such as wrinkles and dynamics. 

To this end, in Section \ref{sec:human_model} we introduce an extension of standard statistical human body models \cite{loper2015smpl} that learns to smoothly diffuse skinning surface parameters, such as rigging weights and blendshape correctives, to any point in 3D space.
In Section \ref{sec:garment_model}, we leverage these learned diffused skinning parameters to define a novel garment deformation model. 
The key idea is to remove the deformations already captured by our diffused body model to built an \textit{unposed} and \textit{deshaped} canonical space of garments. In this space, garments appear in rest pose and mean shape but pose- and shape-dependent wrinkle details are preserved.
In Section \ref{sec:groundtruth_optimization}, we introduce a novel optimization-based strategy to project physics-based simulations to our canonical space of garments. Importantly, we show that the use of the learned diffuse skinning parameters is fundamental for this task, since they enable the correct representation of complex phenomena such as garment-body sliding or loose clothing.

Using projected physics-based simulations as ground-truth data, in Section \ref{sec:generative_space} we describe how we learn a generative space of garment deformations. Key to our success is a novel self-supervised loss enabled by the canonical space of garments, which allows us to exhaustively sample \textit{random} instances of garment deformations (\textit{i.e.}, arbitrary shape, pose, and dynamics for which ground truth data is unavailable) and test collisions against a \textit{constant} body mesh.
Finally, in Section \ref{sec:regressor} we describe a neural network-based regressor that outputs deformed  garments with dynamics, that do not interpenetrate the body, as a function of body shape and motion.

For detailed information about network architectures, training data, parameters, and other implementation details, please refer to the supplementary material.

%% file: method.tex
\section{Canonical Space of Garment Deformations}
The central aim of our method is to obtain a regressor $R$ that infers the deformation of the garment via
\begin{align}
 	\mathbf{X}&=R(\upbeta, \upgamma),
 	\label{eq:regressor_in_model} 
\end{align}
where $\mathbf{X} \in \mathbb{R}^{N_{\text{G}} \times 3}$ is the %
garment deformation in canonical space computed as a function of body shape $\upbeta$ and motion descriptor $\gamma$. 
We will first describe how to obtain the canonical space into which the garment data is transformed, before detailing how the regressor $R$ is trained in Section~\ref{sec:regressor}.

\subsection{Diffused Human Model}
\label{sec:human_model}
Our garment model, defined later in Section~\ref{sec:garment_model}, is driven by a new \textit{diffused} human body model that extends current approaches in order to generalize to vertices beyond the body surface.
More specifically, current body models \cite{feng2015deformable,loper2015smpl,joo2018total} deform a rigged parametric human template
\begin{equation}
	M_\text{B}(\upbeta, \uptheta) = W(T_\text{b}(\upbeta,\uptheta), J(\upbeta), \uptheta,\mathcal{W}) 
	\label{eq:smpl}
\end{equation}
where $W$ is a skinning function (\textit{e.g.}, linear blend skinning, or dual quaternion) with skinning weights $\mathcal{W}$ and pose parameters $\uptheta$ that deforms an unposed parametric body mesh $T_\text{b}(\upbeta,\uptheta)$.  
The nowadays standard SMPL model~\cite{loper2015smpl} defines the unposed body mesh as
\begin{equation}
	T_\text{B}(\upbeta,\uptheta) = \mathbf{T}_\text{b} + B_\text{s}(\upbeta) + B_\text{p}(\uptheta)
\end{equation}
where $\mathbf{T}_\text{b}  \in \mathbb{R}^{N_{\text{B}}\times3}$ is a body mesh template with $N_{\text{B}}$ vertices that is deformed using two blendshapes that output \textit{per-vertex} 3D displacements: $B_\text{s}(\upbeta) \in \mathbb{R}^{N_{\text{B}}\times3}$ models deformations to change the body shape; and $B_\text{p}(\uptheta)  \in \mathbb{R}^{N_{\text{B}}\times3}$ models deformations to correct skinning artifacts. Follow-up works propose additional blendshapes to model soft-tissue \cite{ponsmoll2015dyna,santesteban2020softsmpl} and garments \cite{ma2020dressing3d,alldieck19cvpr,pons2017clothcap}.

We observe that existing data-driven garment models \cite{santesteban2019virtualtryon,patel2020tailor} leverage the human body models defined in Equation \ref{eq:smpl} assuming that clothing closely follows the deformations of the body. Consequently, a common approach is to borrow the skinning weights $\mathcal{W}$ to model the articulation of garments, usually by exhaustively searching the nearest body vertex for each garment vertex in rest pose.
Our key observation is that such naive static assignment cannot correctly model complex nonrigid clothing effects.
The reason is twofold: first, the garment-body nearest vertex assignment must be dynamically updated, for example, when a garment slides over the skin surface; and second, the garment-body vertex assignment cannot be driven only by the closest vertex since this causes undesirable discontinuities in medial-axis areas. 

To address these weaknesses, we propose to extend existing body models formulated in Equation \ref{eq:smpl} by smoothly \textit{diffusing} skinning parameters to any 3D point around the body. 
It is worth mentioning that we are not the first to diffuse surface parameters, but previous works are limited to interpolate inwards to create a volumetric mesh \cite{meekyoung2017physics,romero2020skinmechanics} in a less smooth strategy.
In Section~\ref{sec:garment_model}
we show how our generalization of skinning parameters beyond the body surface is a fundamental piece for our novel garment model.

More formally, we define the functions $\widetilde{\mathcal{W}}(\mathbf{p})$,   $\widetilde{B}_\text{s}(\mathbf{p}, \uptheta)$, and $\widetilde{B}_\text{p}(\mathbf{p}, \uptheta)$ that generalize skinning weights, shape blendshape offset, and pose blendshape offset, respectively, to any point $\mathbf{p} \in \mathbb{R}^3$ by smoothly diffusing the surface values
\begin{align}
	\widetilde{\mathcal{W}}(\mathbf{p}) &= \frac{1}{N} \sum_{\mathbf{q}_n\sim {\mathcal {N}}(\mathbf{p} ,\mathbf{d} )} \mathcal{W}(\phi(\mathbf{q}_n))
\\
	\widetilde{B}_\text{s}(\mathbf{p},\upbeta) &= \frac{1}{N} \sum_{\mathbf{q}_n\sim {\mathcal {N}}(\mathbf{p} ,\mathbf{d} )} {B}_\text{s}(\phi(\mathbf{q}_n),\upbeta)
	\\
	\widetilde{B}_\text{p}(\mathbf{p},\uptheta) &= \frac{1}{N} \sum_{\mathbf{q}_n\sim {\mathcal {N}}(\mathbf{p} ,\mathbf{d} )} {B}_\text{p}(\phi(\mathbf{q}_n),\uptheta)
\end{align}
where $\phi(\mathbf{p})$ computes the closest surface point to  $\mathbf{p} \in \mathbb{R}^3$, $\mathbf{d}$ the distance from $\mathbf{p}$ to the surface body, and ${B}_\text{p}(\mathbf{p},\uptheta)$ a function that returns the 3D offset of the vertex $\mathbf{p}$ computed by the blendshape $B_\text{p}$.
Notice that, for each point, we average the values of $N$ neighbors and therefore mitigate potential discontinuities in areas around a medial-axis.

In order to obtain differentiable functions that seamlessly integrate into an  optimization or learning process, we employ recent works on learning implicit functions and learn $\widetilde{\mathcal{W}}(\mathbf{p})$,   $\widetilde{B}_\text{s}(\mathbf{p}, \upbeta)$, and  $\widetilde{B}_\text{p}(\mathbf{p}, \uptheta)$ 
with fully-connected neural networks.
This additionally yields a very efficient evaluation on modern GPUs.

\subsection{Garment Model}
\label{sec:garment_model}
Our next goal is to define a garment model that is capable of representing the deformations naturally present in real garments, including dynamics, high-frequency wrinkles, and garment-skin sliding. 
To this end, a common approach to ease this task is to decouple the deformations caused by different sources%
, and model each case independently.
For example, Santesteban \textit{et al.} \cite{santesteban2019virtualtryon} decouple deformations due to shape and pose, and Patel \textit{et al.} \cite{patel2020tailor} due to shape, pose, and style. 
More specifically, as discussed in Section \ref{sec:human_model}, both works model pose-dependent deformations leveraging the skinning weights associated with the body in the unposed state and a linear blend skinning technique.
This disentanglement removes many nonlinear deformations and enables to efficiently represent (and learn) deformations due to other sources directly in an \textit{unpose} (\textit{i.e.}, normalized) state. 

We propose to go one step further and remove the shape-dependent deformations already captured by the underlying human body model. This effectively constructs a canonical \textit{unposed} and \textit{deshaped} representation of garments, improving the disentanglement proposed by earlier works. As we show later in Section \ref{sec:generative_space}, this is a fundamental step towards learning a generative space of garment deformations that do not interpenetrate the underlying body.

To formulate our unposed and deshaped garment model we leverage the diffused skinning functions proposed in Section \ref{sec:human_model} 
 \begin{align}
 	M_\text{G}(\mathbf{X}, \upbeta, \uptheta) &=W(T_\text{G}(\mathbf{X}, \upbeta,\uptheta),J(\upbeta), \uptheta,\widetilde{\mathcal{W}}(\mathbf{X})) \label{eq:skinning_variable} 
 \\
 	T_\text{G}(\mathbf{X}, \upbeta,\uptheta) &=\mathbf{X} + \widetilde{B}_\text{s}(\mathbf{X}, \upbeta) + \widetilde{B}_\text{p}(\mathbf{X}, \uptheta),
 	\label{eq:blendshape_variable} 
 \end{align}
where $T_\text{G}()$ is the deformed garment after diffused blendshapes correctives are applied, and $\mathbf{X}$ are the garment deformations in canonical space.
\revised{Notice that our garment model is well-defined for any garment with any topology, thanks to the generalized diffused skinning functions. (\textit{i.e.,} no need to retrain $\widetilde{\mathcal{W}}(),\widetilde{B}_\text{s}(),\widetilde{B}_\text{p}()$ for each garment).}
\revised{The key property of this model is that skinning parameters used to articulate the garment (Equations \ref{eq:skinning_variable} and \ref{eq:blendshape_variable}) are defined as a function of the unposed and deshaped \textit{deformed} garment $\mathbf{X}$.}
This is in contrast to existing methods \cite{santesteban2019virtualtryon,patel2020tailor} that use a \textit{fixed} weight assignment, \revised{usually defined in a relaxed state or template}, and cannot guarantee that the rigging step of the regressed deformed garment does not introduce collisions.

\subsection{Projecting the Ground-truth Data}
\label{sec:groundtruth_optimization}
\begin{figure}
  \includegraphics[width=\linewidth]{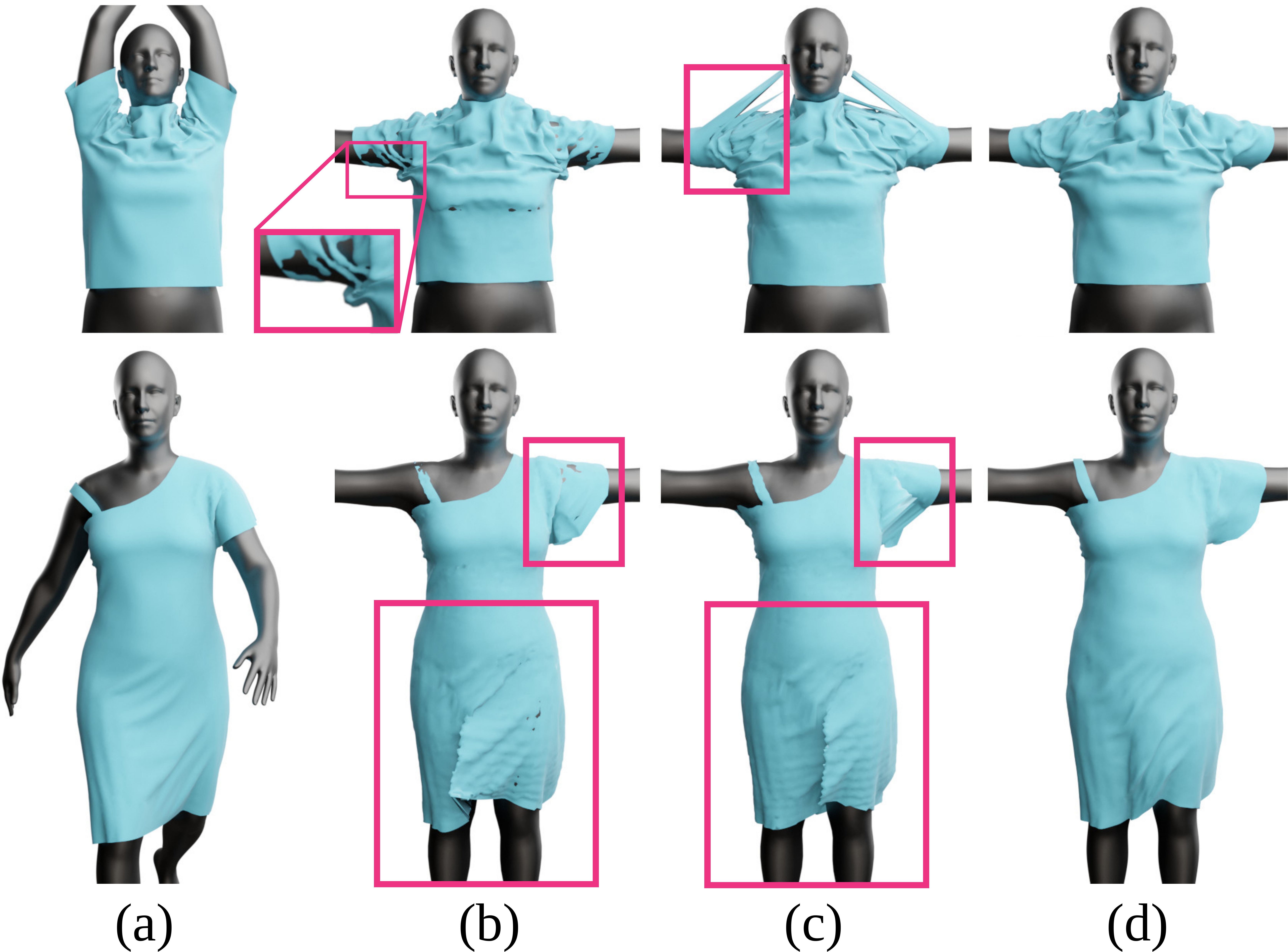}
    \caption{Unposing of a T-shirt and a dress in challenging poses: (a) input mesh; (b) unposing with constant weights \cite{patel2020tailor,santesteban2019virtualtryon}, notice the collisions; (c) unposing with variable weights assigned with nearest vertex, it avoids collisions but introduces skinning artifacts and is not \revised{temporally} stable, better seen in video; (d) unposing with our optimization. %
    }
    \label{fig:comparison_unposing}
\end{figure}

Our ultimate goal is to learn the function $R()$ from Equation \ref{eq:regressor_in_model}, which predicts garment deformations in canonical space, in a data-driven manner.
However, obtaining ground truth data is not trivial since we need to project deformed 3D garments --computed with a physics-based simulator \cite{narain2012arcsim}-- to the unposed and deshaped space.
Previous methods 
formulate the projection to the unposed state as the inverse of the linear blend skinning operation \cite{patel2020tailor,santesteban2019virtualtryon,pons2017clothcap}. Due to their static rigging weights assignment, this operation can introduce body-garment collisions in the unposed state for frames where the garment has deformed significantly or slid in the tangential direction of the body (see Figure \ref{fig:comparison_unposing}b).
Even if a data-driven method can potentially learn to fix these artifacts to output collision-free \textit{posed} deformations, our key contribution discussed in detail in Section \ref{sec:generative_space} is to show that if a collision-free projection-and-unprojection operation exists, then the learning can be defined entirely in the unposed and deshaped state. This carries many positive properties that we discuss later.

We therefore need an strategy to project ground-truth garments to our canonical space, without introducing collisions.
Notice that
we cannot use the inverse of Equation \ref{eq:skinning_variable} because the diffused skinning $\widetilde{\mathcal{W}}(\mathbf{X})$ are only defined for unposed shapes.
Furthermore, exhaustive search of garment-body nearest vertices for each frame \revised{is highly} expensive and introduces discontinuities in medial axis areas (see Figure \ref{fig:comparison_unposing}c).
Therefore, we propose a new optimization-based strategy 
to find the optimal vertex positions of the garment in the canonical space.
Formally, given a ground-truth deformed garment mesh $\text{M}_\text{G}$ (\textit{i.e.,} generated with physics-based simulation)  with known pose $\uptheta$ and shape $\upbeta$, we find its unposed and deshaped representation $\mathbf{X}$ by minimizing
\begin{equation}
    \min_{\mathbf{X}} \quad \mathcal{E}_\text{rec} + \omega_1\mathcal{E}_\text{strain} + \omega_2\mathcal{E}_\text{collision}.
    \label{eq:minimization}
\end{equation}
In the minimization objective, 
the data term %
\begin{align}
	\mathcal{E}_\text{rec} &= \begin{Vmatrix}\text{M}_\text{G} - M_\text{G}(\mathbf{X}, \upbeta, \uptheta)\end{Vmatrix}_2^2
\end{align}
aims at reducing the difference between the simulated garment, and the unposed and deshaped representation projected back to the original state.
Notice that $M_\text{G}(\mathbf{X}, \upbeta, \uptheta)$, defined in Equation \ref{eq:skinning_variable}, is well defined for any set of 3D vertices $\mathbf{X}$, and it is fully differentiable thanks to the diffused skinning parameters. 

The regularization term %
{
\medmuskip=1mu
\thinmuskip=1mu
\thickmuskip=1mu
\begin{align}
	\mathcal{E}_\text{strain} &= \begin{Vmatrix} \frac{1}{2}(F(\mathbf{T_\text{G}(\mathbf{X}, \upbeta, \uptheta)})^\top  F(\mathbf{T_\text{G}(\mathbf{X}, \upbeta, \uptheta)}) - \text{I})  \end{Vmatrix}_2^2
\end{align}
}
penalizes unrealistic deformations. To measure the amount of deformation of each triangle we use the Green-Lagrange strain tensor, which is rotation and translation invariant. $F$ denotes the deformation gradient of each triangle.

Lastly, %
we include a term to prevent optimized vertex positions $\mathbf{X}$ to interpenetrate with the underlying body:
\begin{equation}
	\mathcal{E}_\text{collision} = \text{max}(\upepsilon - SDF(\mathbf{X}), 0)
\end{equation}
This term requires to compute the distance to the body surface for all vertices of the deformed garment, which is usually modeled with a Signed Distance Field (SDF). 
We leverage the fact that bodies in our canonical space are represented with a \textit{constant body mesh}, and therefore the SDF is static and can be precomputed. In practice, and inspired by recent works on implicit function learning \cite{park2019deepsdf,atzmon2020sal,chen2019learning,sitzmann2020siren}, we learn the SDF with a shallow fully-connected network that naturally provides a fully differentiable formulation.

To optimize a sequence, we initialize the optimization with the result of the previous frame. 
This not only accelerates convergence, but also contributes to stabilize the projection over time.
For the first frame, we initialize the optimization with the garment template, which is obtained by simulating the garment with the average body model (\textit{i.e.}, $\uptheta$ and $\upbeta$ set to zero).

\subsection{Generative Garment Deformation Subspace}
\label{sec:generative_space}
With the garment model defined in Section \ref{sec:garment_model}, and the strategy to project ground truth data into our canonical space  defined in Section \ref{sec:groundtruth_optimization}, we could train a data-driven method (\textit{e.g.} a neural network) to learn the garment deformation regressor $R()$ defined in Equation \ref{eq:regressor_in_model}.
However, even though our garment model is designed in such a way that the (un)projection operation between canonical space and posed space does not introduce collisions, residual errors in the optimization of the regressor $R( )$ could lead to regressed deformed garments $\mathbf{X}$ with body-garment collisions in the canonical space, which would inevitably propagate to the posed space.
In fact, this is a common source of collisions in all data-driven methods \cite{gundogdu2019garnet,patel2020tailor,santesteban2019virtualtryon,wang2018multimodalspace}.

Our key contribution to address this challenge is to learn a compact subspace for garment deformations that \textit{reliably solves} garment-body interpretations. To do so, we leverage the fact that in our unposed and deshaped canonical representation of garments, the underlying body shape is \textit{constant}, namely, it is a body shape with $\upbeta=\mathbf{0}$ and $\uptheta=\mathbf{0}$.
This property enables us to train a variational autoencoder (VAE) to learn a \textit{generative} space of garment deformations with a novel self-supervised collision loss term that is independent of the underlaying body and shape, and therefore naturally generalizes to arbitrary bodies. More specifically, we train the VAE with a loss
\begin{equation}
    \mathcal{L}_\text{VAE} = \mathcal{L}_\text{rec} + \lambda_1\mathcal{L}_\text{laplacian} + \lambda_2\mathcal{L}_\text{collision} +
    \lambda_3\mathcal{L}_\text{KL}.
\end{equation}
We define the standard VAE reconstruction term as
\begin{equation}
    \mathcal{L}_\text{rec} = \begin{Vmatrix}\mathbf{X} - D(E(\mathbf{X}))\end{Vmatrix}_1,
\end{equation}
where $E()$ and $D()$ are the encoder and decoder networks, respectively. Since $\mathcal{L}_\text{rec}$ does not take into account the neighborhood of the vertex, we add an additional loss term that penalizes error between the mesh laplacians \cite{taubin95laplacian,wang2019intrisicspace}
\begin{equation}
    \mathcal{L}_\text{laplacian} = \begin{Vmatrix}\Delta\mathbf{x} - \Delta D(E(\mathbf{X}))\end{Vmatrix}_1
\end{equation}

To enforce a subspace free of garment-body collisions, we propose the collision term
\begin{equation}
\begin{multlined}
\mathcal{L}_\text{collision} = max(\upepsilon - SDF(D(E(\mathbf{X})), 0)\\
+ max(\upepsilon - SDF(D(\mathbf{\bar{X}}_\text{rand})), 0)
\end{multlined}
\label{eq:collision_loss}
\end{equation}
where $\mathbf{\bar{X}}_\text{rand} \sim \mathcal{N}(0, 1)$. The first term penalizes collisions in the reconstruction of train data.
Our fundamental contribution is the second term, $max(\upepsilon - SDF(D(\mathbf{\bar{X}}_\text{rand})), 0)$, that samples the latent space and, enabled by the deshaped and unpose canonical representation, checks collisions against a \textit{constant body mesh} with a self-supervised strategy (\textit{i.e.}, we do not need ground truth garments for this term).
This key ingredient allows us to exhaustively sample the latent space and learn a compact garment representation that reliably solves garment-body interpenetrations.
As already highlighted, since our garment model is designed to not to introduce body-garment collisions in both the projection and unprojection operations, garment deformations regressed in the generative subspace do not suffer from collisions even in unseen (\textit{i.e.,} test) sequences.  

The self-supervised loss is only useful if the values are sampled from the same distribution as the data. For this purpose, we include an additional term $\mathcal{L}_\text{KL}$ to enforce a normal distribution in our latent space.

\begin{figure}
	\centering
	\includegraphics[width=\linewidth, trim=0pt 280pt 0pt 280pt,clip]{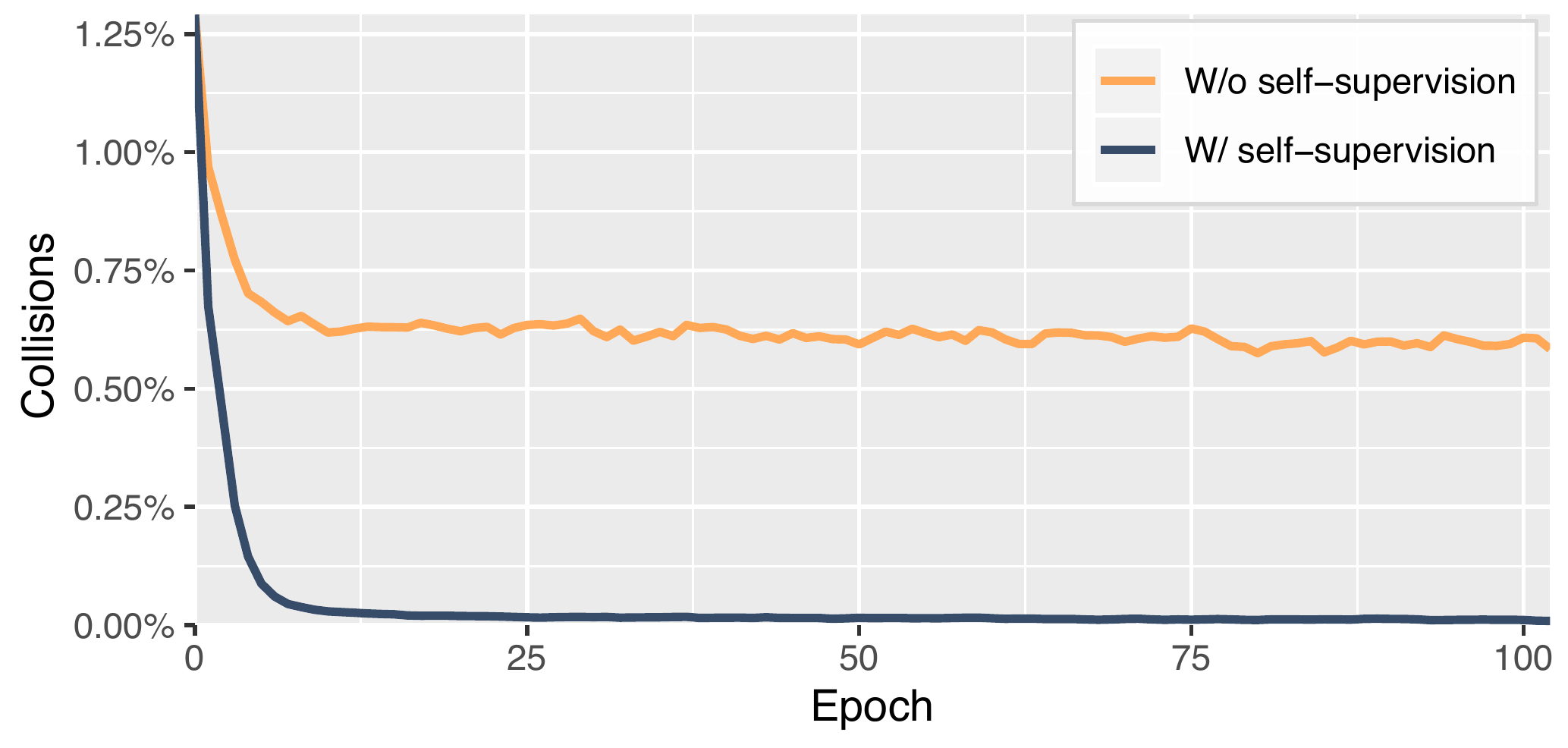}
	\caption{Number of body-garment collisions, evaluated in a test set, during the training of the generative subspace. Our novel self-supervised term, described in Equation \ref{eq:collision_loss}, is the key term to reduce collisions in unseen sequences.
	}
	\label{fig:collision_loss_training}
\end{figure}

%% file: regressor.tex
\section{Regressing Garment Deformations}
\label{sec:regressor}
Once we have built our generative garment subspace, we encode the ground-truth data and use it to train the recurrent regressor $R(\upbeta,\gamma)$ from Equation \ref{eq:regressor_in_model}, which predicts garment deformations as a function of body shape $\upbeta$ and motion $\gamma$.

Our motion descriptor $\upgamma$ carries information of the current pose as well as its global movement. The off-the-shelf encoding for pose information is to use the joint rotations $\uptheta \in \mathbb{R}^{72}$  of the underlying human model, but this representation suffers from several problems such as discontinuities, redundant joints, and unnecessary degrees of freedom.
Instead, we adopt a more compact, learned pose descriptor $\bar{\uptheta} \in \mathbb{R}^{10}$ \cite{santesteban2020softsmpl} which we found to generalize better.
We build the motion vector $\upgamma$ for a given frame by concatenating the  descriptor to the velocities and accelerations (computed with finite differences) of the pose, the global rotation $\textbf{K}$ (represented as Euler angles) and translation $\textbf{H}$ 
\begin{equation}
\upgamma = \{\bar{\uptheta}, \dv{\bar{\uptheta}}{t},\dv[2]{\bar{\uptheta}}{t}, \dv{\text{H}}{t}, \dv[2]{\text{H}}{t}, \dv{\text{K}}{t}, \dv[2]{\text{K}}{t}\}
\end{equation}

The regressor takes as input the motion descriptor $\upgamma \in \mathbb{R}^{42}$ and the shape coefficients $\upbeta \in \mathbb{R}^{10}$ and predicts the encoded garment deformation $\mathbf{\bar{X}}_\text{pred} \in \mathbb{R}^{25}$. 
To learn dynamic effects that depend on previous frames, we use Gated Recurrent Units \cite{cho2014gru} as the building blocks of our model.
We train using the L1-error of encoded canonical space positions, velocities, and accelerations, which we find improves dynamics compared to optimizing positions alone.
\begin{equation}
    \mathcal{L}_\text{R} = \mathcal{L}_\text{pos} + \rho_1\mathcal{L}_\text{vel} + \rho_2\mathcal{L}_\text{acc}
\end{equation}

%% file: evaluation.tex
\section{Evaluation}
\subsection{Quantitative Evaluation}
To quantitatively evaluate the ability of our compact generative subspace to solve body-garment collisions, we show the number of collisions during the training in Figure \ref{fig:collision_loss_training}, evaluated on a \textit{test set} that includes 4 unseen sequences and 17 different shapes. 
Specifically, we plot an ablation study that shows, in orange, the collisions remaining at each epoch when using only the supervised collision loss (\textit{i.e.}, 1st term of Equation \ref{eq:collision_loss}), and, in black, when also using our self-supervision with the 2nd term of Equation \ref{eq:collision_loss}. 
The latter dramatically improves the collision handling, and it shows the generalization capabilities of our approach by reaching values close to 0 collisions in unseen sequences.

\begin{figure}
	\centering
	\vspace{-0.1cm}
	\begin{tikzpicture}
		\node[] (image) at (0,0) 
		{\includegraphics[width=\linewidth, trim=15pt 280pt 8pt 285pt,clip]{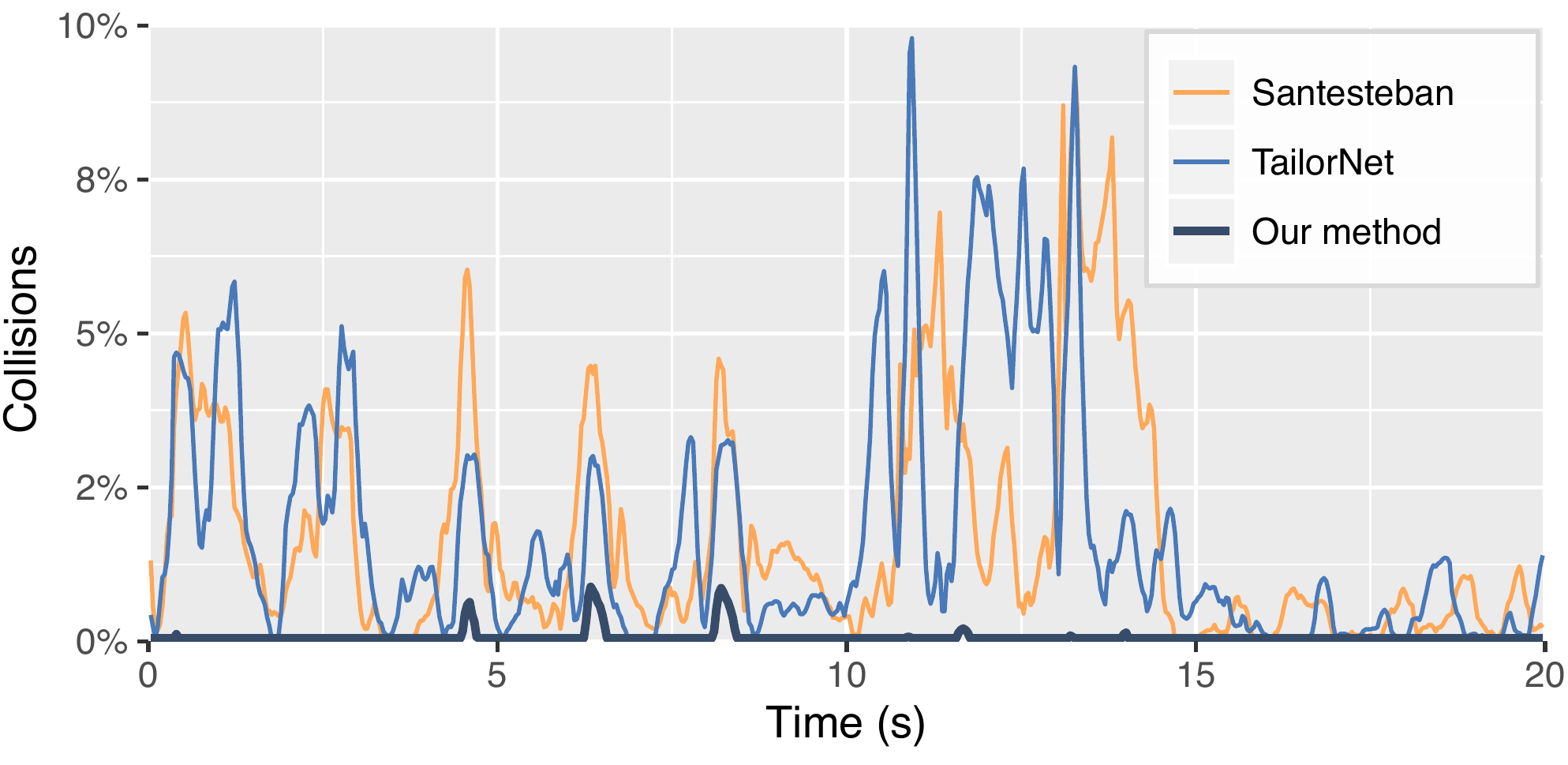}};
		\node[right of=image,xshift=2.48cm, yshift=1.160cm, scale=0.6] (textA) {\cite{patel2020tailor}};
		\node[right of=image,xshift=2.8cm, yshift=1.55cm, scale=0.6] (textB) {\cite{santesteban2019virtualtryon}};
	\end{tikzpicture}
\vspace{-0.6cm}
	\caption{Quantitative evaluation of collisions per frame in test sequence 86\_07. See supplementary video for a qualitative visualization of this comparison.
	}

	\label{fig:collisions_sequence}
\end{figure}

In Figure \ref{fig:collisions_sequence} we show a quantitative evaluation of the collisions in a test sequence from the AMASS dataset \cite{AMASS:ICCV:2019}, and compare our results with the recent works of Santesteban \textit{et al.} \cite{santesteban2019virtualtryon} and TailorNet \cite{patel2020tailor}.
These previous methods, without the postprocessing step, generate garment deformations that consistently collide with the underlying body mesh. In contrast, our method directly regresses garment deformations with almost no collisions. Importantly, the primary source of the remaining collisions for our method are self-intersections in the body mesh already present AMASS dataset (\textit{e.g.}, a hand interpenetrates the torso).

Furthermore, in Table \ref{table:collisions_quantitative} we present an ablation study evaluated on test sequences from AMASS \cite{AMASS:ICCV:2019} with 53,998 frames and 20 body shapes.
We report the number of collisions for 3 configurations of our method: without the full collision loss, without the self-supervised term, and the full model. 
All components of our model contribute, leading to a residual of 0.09\% with our full model. In contrast, competitive methods suffer from a significantly higher number. %

\begin{table}[]
	\setlength{\tabcolsep}{5.5pt}
	\begin{tabular}{lccccc}
		& 
		\rot{\vtop{\hbox{\strut TailorNet \cite{patel2020tailor}}\hbox{\strut \footnotesize	{(w/o postprocess)}}}}  
		&  
		\rot{\vtop{\hbox{\strut Santesteban \cite{santesteban2019virtualtryon}}\hbox{\strut \footnotesize	{(w/o postprocess)}}}} 
		&\rot{\vtop{\hbox{\strut Our method}\hbox{\strut \footnotesize	{(w/o collision loss)}}}} 
		&
		\rot{\vtop{\hbox{\strut Our method}\hbox{\strut \footnotesize	{(w/o self-supervision)}}}}
		&
		\rot{Our method} \\
		\midrule
		Collisions & 5.70\% & 8.80\% & 0.62\% & 0.24\% & \textbf{0.09}\% \\
		\midrule
	\end{tabular}
	\vspace{-0.2cm}
	\caption{
		Average number of collisions in 105 test motions from the AMASS dataset \cite{AMASS:ICCV:2019}. 
	}
	\label{table:collisions_quantitative}
\end{table}

{
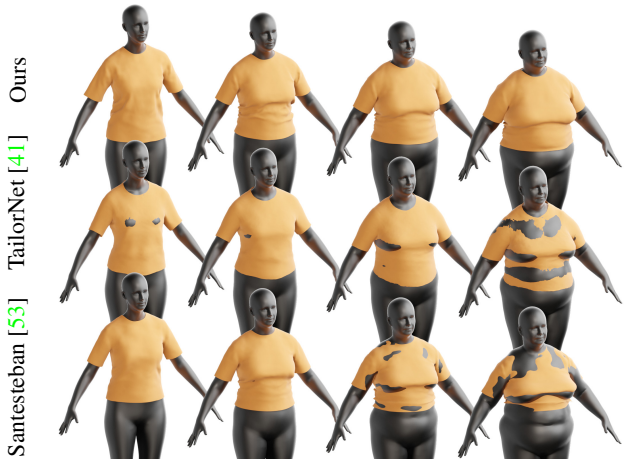
\begin{figure}[t]
	\centering
		\begin{tikzpicture}
			\node[rotate=90] (textA) at (-1,0) {\small{\quad\quad\quad TailorNet \cite{patel2020tailor}}};
			\node[rotate=90, right of=textA, xshift=-2.85cm] (textB){\small{Santesteban \cite{santesteban2019virtualtryon}}};
			\node[rotate=90, right of=textA, xshift=1.1cm] (textC){\small{Ours}};
			\node[right of=textA, xshift=3.3cm,yshift=0.1cm] (image) 
			{\includegraphics[width=0.43\textwidth,trim=10pt 0 0 0, clip]{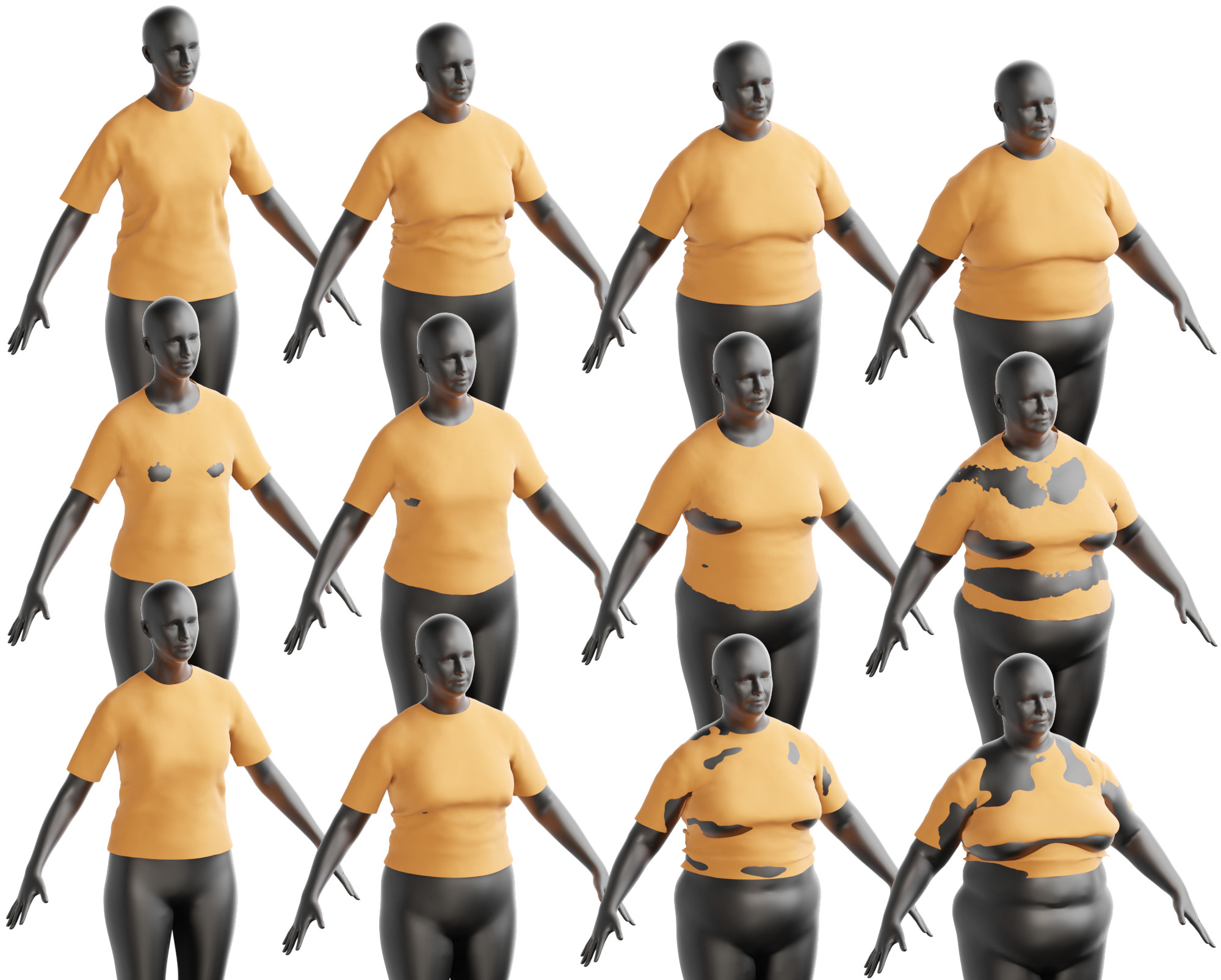}};
		\end{tikzpicture}
    \caption{Generalization to new shapes. Interpolation between two unseen body shapes (left and right) from the AMASS dataset \cite{AMASS:ICCV:2019}. Our deshaped canonical space avoids collisions even in shapes far from the training data.    %
    }
	\label{fig:generalization_shape}
\end{figure}
}

\begin{figure}

  \begin{subfigure}{0.32\linewidth}
    \includegraphics[width=\linewidth]{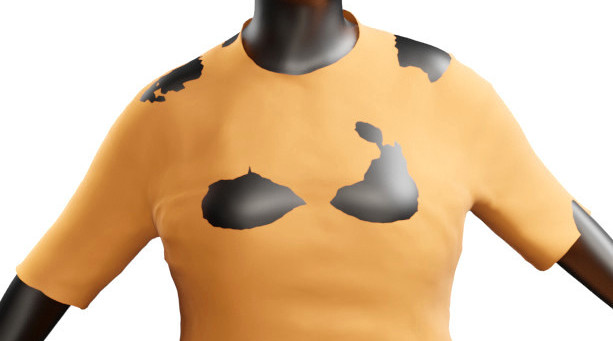}
    \caption{TailorNet \cite{patel2020tailor}\\\footnotesize{(w/o postprocess)}} \label{fig:1a}
  \end{subfigure}%
  \hspace*{\fill}   %
  \begin{subfigure}{0.32\linewidth}
    \includegraphics[width=\linewidth]{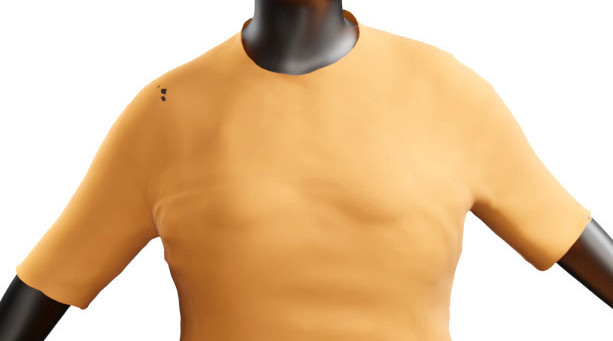}
    \caption{TailorNet \cite{patel2020tailor}\\\footnotesize{\quad(w/ postprocess)}} \label{fig:1b}
  \end{subfigure}%
  \hspace*{\fill}   %
  \begin{subfigure}{0.32\linewidth}
    \includegraphics[width=\linewidth]{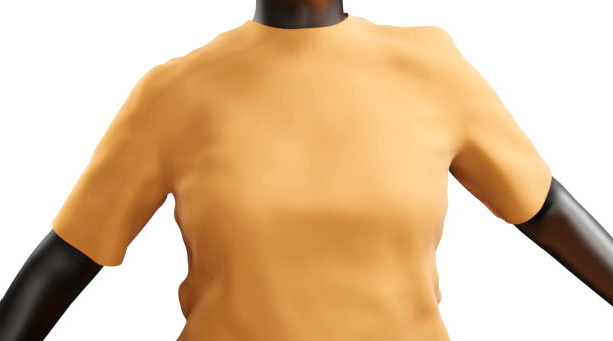}
    \caption{Our method\\ \quad} \label{fig:1c}
  \end{subfigure}
  \caption{Fixing collisions in a postprocess step can introduce undesired bulges, see chest area in (b).} 
\label{fig:comparison_collisions_tailornet}
  
\end{figure}

\subsection{Qualitative Evaluation}
We qualitatively evaluate the output of our method and compare to recent approaches.
Please notice that the visual quality of our results is much better demonstrated in the supplementary video, which contains an exhaustive qualitative analysis, including a real-time demo in test sequences with unseen shapes from the AMASS dataset \cite{AMASS:ICCV:2019}.
 
In Figure \ref{fig:generalization_shape} we qualitatively evaluate the generalization capabilities our method to \textit{unseen body shapes}. 
Specifically, we interpolate between 2 extremely different real shapes from AMASS \cite{AMASS:ICCV:2019}, and compare to state-of-the-art data-driven garment models.
Importantly, the input shapes are far beyond the range of our training data, therefore here we are also evaluating the \textit{extrapolation} capabilities of the methods.
Our method handles such extremely challenging cases very well and does not show visible garment-body collision, while existing methods \cite{santesteban2019virtualtryon,patel2020tailor} suffer from very noticeable interpenetrations. In Figure \ref{fig:comparison_collisions_tailornet} we show that, although a postprocessing step can effectively mitigate this issue, it can also introduce additional problems.

\begin{figure}
    \centering
    \includegraphics[width=\linewidth]{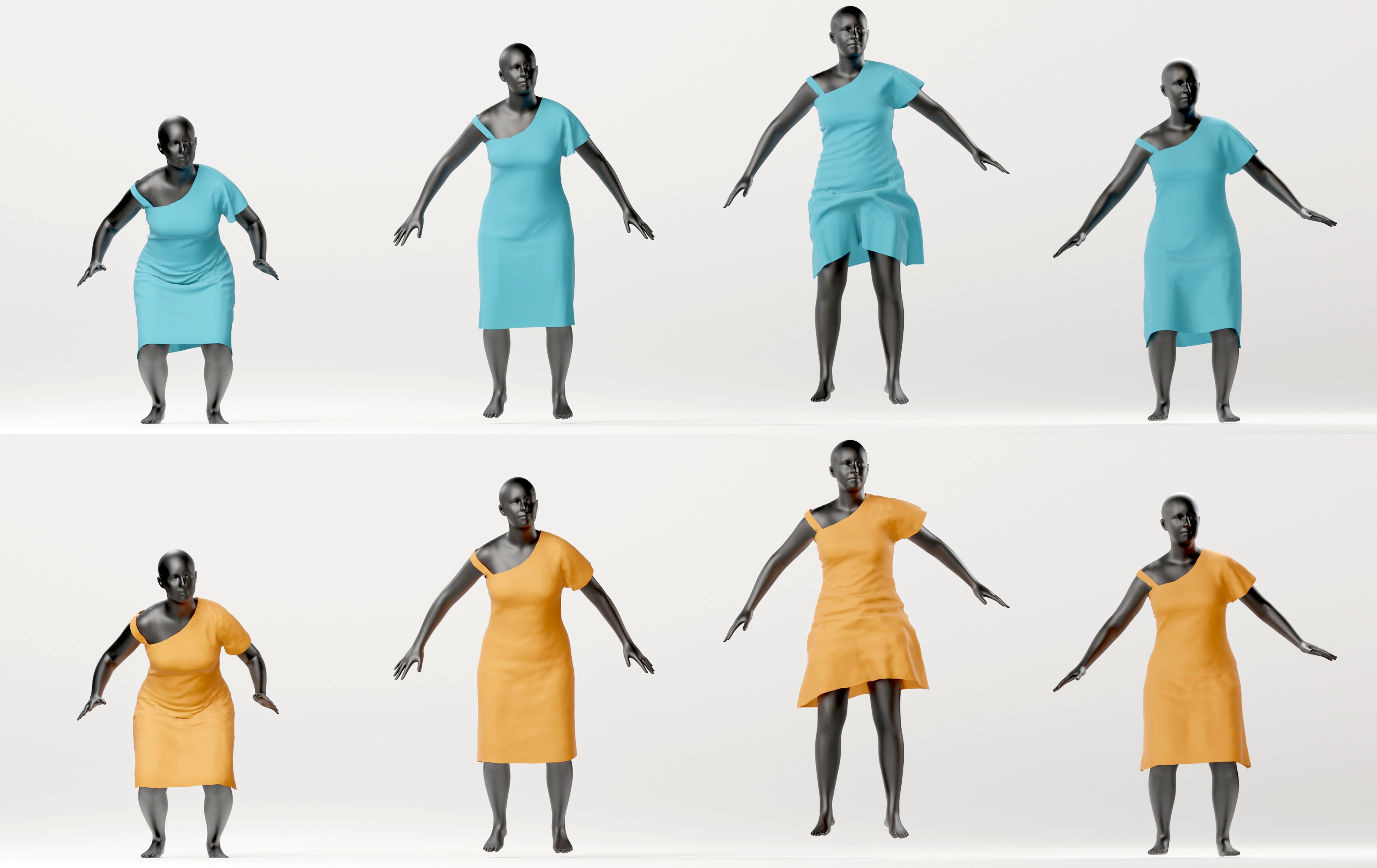}
    \caption{Generalization to new motions. Qualitative comparison with physical simulation \cite{narain2012arcsim} (top) in sequence 01\_01. Our model (bottom) synthesizes highly realistic dynamics and wrinkles even for challenging unseen motions. 
     }
    \label{fig:generalization_motion}
\end{figure}

In Figure \ref{fig:generalization_motion} we evaluate the generalization capabilities of our approach to \textit{unseen motions}, and we compare our results against those of a physics-based simulator.
Notice that our method is the first to showcase such a highly-challenging scenario featuring a dress sequence with dynamics.
\subsection{Runtime Performance}
In Table \ref{table:runtime} we show our runtime performance in a regular desktop PC (AMD Ryzen 7 2700 CPU, Nvidia GTX 1080 Ti GPU, and 32GB of RAM). 
We produce detailed meshes at high frame rates, even for garments with many triangles.
\begin{table}
	\centering
	\begin{tabular}{rrccc} 
		\toprule 
		& Triangles & Regressor & Decoder & Projection \\
		\midrule
		T-shirt & 8,710 & 1.7 ms & 1.6 ms & 1.4 ms \\
		
		Dress & 23,949 & 1.7 ms & 3.5 ms & 2.9 ms\\
		\bottomrule
	\end{tabular}
	\caption{Execution time of each step of our model.}
	\label{table:runtime}
\end{table}

%% file: conclusions.tex
\section{Conclusions and Future Work}
We have presented a first algorithm to learn garment deformations such that they are essentially collision free. We achieve this by a diffused, volumetric representation of the underlying body together with the construction of a subspace for the garment that yields a differentiable, canonical configuration. 
This subspace is crucial for the regression of the garment deformation and its dynamics.
Our algorithm not only allows for avoiding collisions, it also reduces complexity for inference, such that a learned representation yields higher quality than previously achievable.
Generated garments exhibit a large amount of spatial and temporal detail, and can be inferred extremely quickly.

Additionally, the high quality of the garments generated by our method and the differentiable nature of our algorithm point to a large number of highly interesting avenues for follow up work: from virtual try-on applications  \cite{han2018viton}, to inverse problems in computer vision \cite{lassner2017generative}. Extending our method to handle collisions between multiple layers of clothing is also a promising line of future research.

\noindent\paragraph{Acknowledgments.}Igor Santesteban was supported by the Predoctoral Training Programme of the Basque Government (PRE\_2020\_2\_0133). The work was also funded in part by the European
Research Council (ERC Consolidator Grant no. 772738 TouchDesign) and Spanish Ministry of Science (RTI2018-098694-B-I00
VizLearning).